\documentclass[10pt,twocolumn,letterpaper]{article}

\usepackage{cvpr}
\usepackage{times}
\usepackage{epsfig}
\usepackage{graphicx}
\usepackage{amsmath}
\usepackage{amssymb}

\usepackage{mmstyles}
\usepackage{color}
\usepackage{array}
\usepackage{multirow}

\newcolumntype{R}[1]{>{\raggedright\arraybackslash}p{#1}}
\newcolumntype{L}[1]{>{\raggedleft\arraybackslash}p{#1}}
\newcolumntype{P}[1]{>{\centering\arraybackslash}p{#1}}
\newcolumntype{M}[1]{>{\centering\arraybackslash}m{#1}}

\makeatletter
\def\hlinew#1{%
  \noalign{\ifnum0=`}\fi\hrule \@height #1 \futurelet
   \reserved@a\@xhline}
\makeatother

\usepackage[breaklinks=true,bookmarks=false]{hyperref}

\cvprfinalcopy % *** Uncomment this line for the final submission

\def\cvprPaperID{462} % *** Enter the CVPR Paper ID here

\begin{document}

%%%%%%%%% TITLE
\title{Semantics Disentangling for Text-to-Image Generation}

\author{Guojun Yin$^{1, 2}$, Bin Liu$^1$, Lu Sheng$^{2, 4}\thanks{Lu Sheng is the corresponding author.}$, Nenghai Yu$^1$, Xiaogang Wang$^2$, Jing Shao$^3$ \\
\normalsize $^1$University of Science and Technology of China, Key Laboratory of  Electromagnetic Space Information, \\
\normalsize The Chinese Academy of Sciences,  $^2$ CUHK-SenseTime Joint Lab, The Chinese University of Hong Kong\\
\normalsize $^3$SenseTime Research, $^4$College of Software, Beihang University\\
\tt\small gjyin@mail.ustc.edu.cn, lsheng@buaa.edu.cn, \\
\tt\small \{flowice,ynh\}@ustc.edu.cn, xgwang@ee.cuhk.edu.hk, shaojing@sensetime.com
}

\maketitle
%\thispagestyle{empty}

%%%%%%%%% ABSTRACT
\begin{abstract}

Synthesizing photo-realistic images from text descriptions is a challenging problem.
Previous studies have shown remarkable progresses on visual quality of the generated images.
In this paper, we consider semantics from the input text descriptions in helping render photo-realistic images. 
However, diverse linguistic expressions pose challenges in extracting consistent semantics even they depict the same thing. 
To this end, we propose a novel photo-realistic text-to-image generation model that implicitly disentangles semantics to both fulfill the high-level semantic consistency and low-level semantic diversity. 
To be specific, we design (1) a Siamese mechanism in the discriminator to learn consistent high-level semantics, and (2) a visual-semantic embedding strategy by semantic-conditioned batch normalization to find diverse low-level semantics. 
Extensive experiments and ablation studies on CUB and MS-COCO datasets demonstrate the superiority of the proposed method in comparison to state-of-the-art methods.

\end{abstract}

%%%%%%%%% BODY TEXT
\section{Introduction}
\label{sec:intro}

The rapid progress of the Generative Adversarial Networks (GAN)~\cite{goodfellow2014gan,li2017perceptual,bousmalis2017unsupervised,kaneko2017generative} brings a remarkable evolution in natural image generation with diverse conditions.
In contrast to conditions such as random noises, label maps or sketches, it is a more natural but challenging way to generate an image from a linguistic description (text) since (1) the linguistic description is a natural and convenient medium for a human being to describe an image, but (2) cross-modal text-to-image generation is still challenging.

%=========== fig: fig1 ==================
\begin{figure}[t]
\centering
\includegraphics[width=0.95\linewidth]{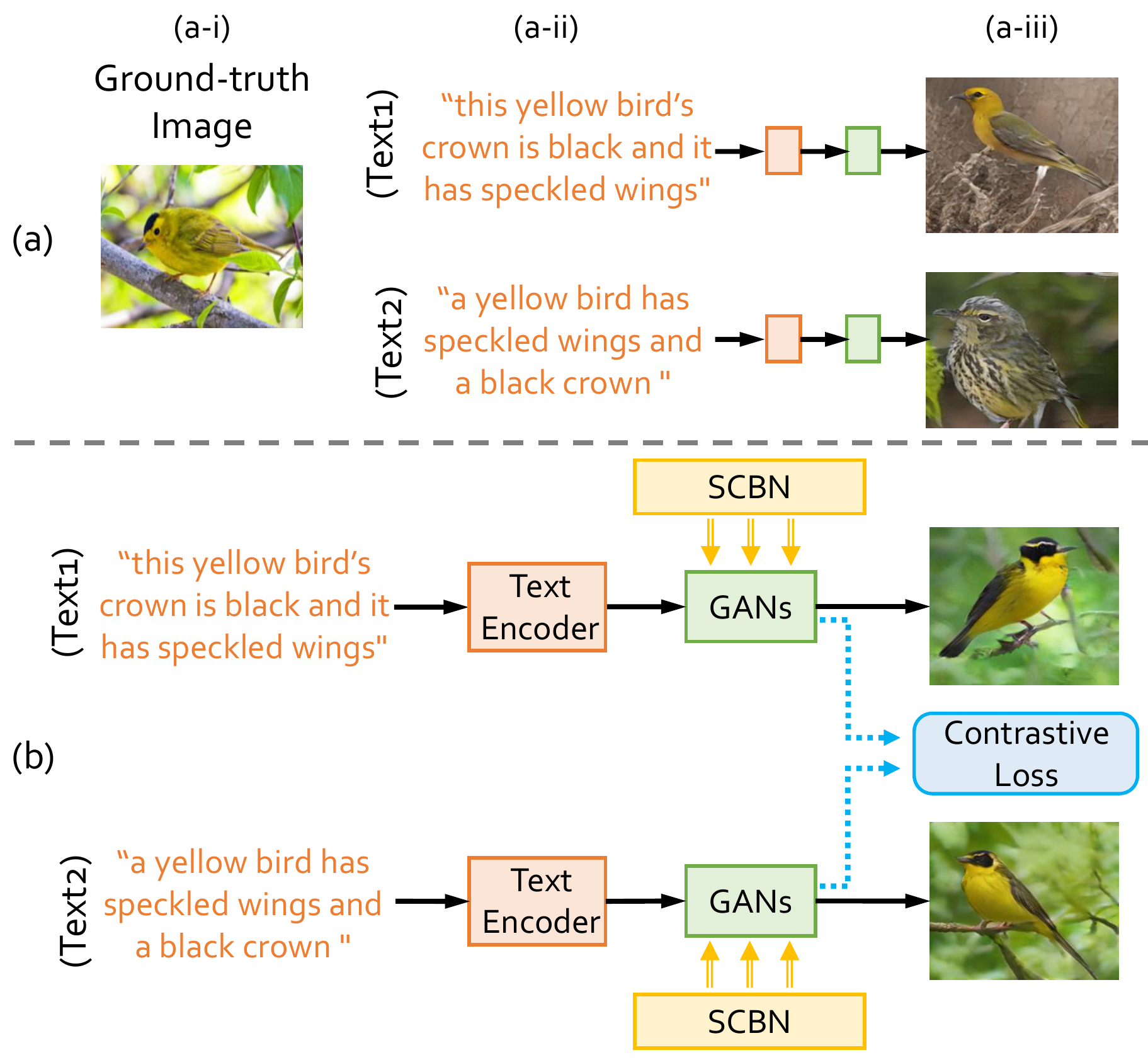}
% % \vspace{-0.5cm}
\caption{ 
Given the language descriptions in (a-ii), their corresponding images are generated by existing GANs in (a-iii). Compared to the groundtruth image in (a-i), such holistic subjective text may lead generation deviation (a-iii) due to the lacking of common and distinct semantic meanings. The proposed SD-GAN in (b) distills the semantic commons by a Siamese structure and retains semantic diversities \& details via a semantic-conditioned batch normalization.
}
\label{fig:fig1}
% \vspace{-0.35cm}
\end{figure}
%=========== fig: fig1 ==================

Existing text-to-image generation works~\cite{zhang2017stackgan,xu2017attngan,zhang2018photographic,hong2018inferring,reed2016generative} mainly focus on increasing the visual quality and resolution of the generated images by either a stacked coarse-to-fine generator structure~\cite{zhang2017stackgan,hong2018inferring} or an attention-guided generation procedure~\cite{xu2017attngan}.
However, these methods neglect one important phenomenon that the human descriptions for a same image are highly subjective and diverse in their expressions, it means that naively using these texts as unique descriptions to generate images would often produce unstable appearance patterns that are far apart from the ground-truth images.
For example, when given different descriptions (Fig~\ref{fig:fig1}(a-ii)) for the same ground-truth image in Fig.~\ref{fig:fig1}(a-i), the generated images in Fig.~\ref{fig:fig1}(a-iii) by~\cite{xu2017attngan} present various appearance patterns apart from the groundtruth, not even similar to the same kind of bird.
It shows that the rich variations of linguistic expressions pose challenges in extracting consistent semantic commons from different descriptions of the same image. Variations of descriptions may lead to deviated image generation even if they describe the same bird with very similar semantic expressions.

To address this issue, in this paper, we propose a novel photo-realistic text-to-image generation method that effectively exploit the semantics of the input text within the generation procedure, named as \emph{Semantics Disentangling Generative Adversarial Network} (SD-GAN). 
The proposed SD-GAN distills the \textit{semantic commons} from texts for image generation consistency and meanwhile retains the \textit{semantic diversities \& details} for fine-grained image generation.

Inspired by the advantages of Siamese structure used in different tasks~\cite{varior2016gated,varior2016siamese,chung2017two,ge2018fdgan,Zhu_2018_ECCV} which can find the similarity between a pair of sequences, 
we treat our discriminator as an image comparator so as to preserve the semantic consistency among the generated images as long as their descriptions are comprehensive and refer to the same semantic contents. 
Specifically, the proposed SD-GAN uses a Siamese scheme with a pair of texts as input and trained with the contrastive loss shown in Fig.~\ref{fig:fig1}(b).
Denote \textit{intra-class} pair as the same groundtruth image with different descriptions while \textit{inter-class} pair as the different groundtruth image with different descriptions. 
By the SD-GAN, the \textit{intra-class} pairs with similar linguistic semantics should generate consistent images that have smaller distances in the feature space of the discriminator, while \textit{inter-class} pairs have to bear much larger distances.
Since we do not have text-to-semantic embedding structure before our image generator, this special training strategy also forces the text-to-image generator has an inherent distillation of semantic commons from diverse linguistic expressions.

To some extent, the Siamese structure indeed distills the \textit{semantic commons} from texts but meanwhile ignores the \textit{semantic diversities \& details} of these descriptions even from the same image.
To maintain the semantic diversities from the texts, the detailed  linguistic cues are supposed to be embedded into visual generation.
Previous works try to guide visual generation by taking the text features as the input to the generator~\cite{zhang2017stackgan,zhang2017stackgan++,xu2017attngan}.
From another perspective, we reformulate the batch normalization layer within the generator, denoted as \textit{Semantic-Conditioned Batch Normalization} (SCBN) in Fig.~\ref{fig:fig1}(b). The proposed SCBN enables the detailed and fine-grained linguistic embedding to manipulate the visual feature maps in the generative networks.

Our contributions are summarized as follows:

\noindent 1) \textit{Distill Semantic Commons from Text-} The proposed SD-GAN distills semantic commons from the linguistic descriptions, based on which the generated images can keep generation consistency under expression variants. To our best knowledge, it is the first time to introduce the Siamese mechanism into the cross-modality generation.

\noindent 2) \textit{Retain Semantic Diversities \& Details from Text-} To complement the Siamese mechanism that may lose unique semantic diversities, we design an enhanced visual-semantic embedding method by reformulating the batch normalization layer with the instance linguistic cues. 
The linguistic embedding can further guide the visual pattern synthesis for fine-grained image generation.

\noindent 3) The proposed SD-GAN achieves the state-of-the-art performance on the CUB-200 bird dataset~\cite{wah2011cub200} and MS-COCO dataset~\cite{lin2014microsoftcoco} for text-to-image generation.

%-------------------------------------------------------------------------
\section{Related Works}
\label{sec:related_works}
% \vspace{-0.15cm}
% % % \vspace{0.1cm}
\noindent\textbf{Generative Adversarial Network (GAN) for Text-to-Image.}
Goodfellow \etal ~\cite{goodfellow2014gan} first introduced the adversarial process to learn generative models. 
The Generative Adversarial Network (GAN) is generally composed of a generator and a discriminator, where the discriminator attempts to distinguish the generated images from real distribution and the generator learns to fool the discriminator. 
A set of constraints are proposed in previous works ~\cite{radford2015unsupervised,huang2017stacked,nowozin2016f,Ge_2018_ECCV,Wu_2018_ECCV} to improve the training process of GANs, \eg, interpretable representations are learned by using additional latent code in~\cite{chen2016infogan}.
GAN-based algorithms show excellent performance in image generation~\cite{li2017perceptual,bousmalis2017unsupervised,kaneko2017generative,nguyen2017shadow,Wang_2018_ECCV,brock2018large,Lu_2018_ECCV}.
Reed \etal ~\cite{reed2016learning} first showed that the conditional GAN was capable of synthesizing plausible images from text descriptions. Zhang \etal \cite{zhang2017stackgan,zhang2017stackgan++} stacked several GANs for text-to-image synthesis and used different GANs to generate images of different sizes.
Their following works~\cite{zhang2018photographic,xu2017attngan} also demonstrated the effectiveness of stacked structures for image generation. Xu \etal \cite{xu2017attngan} developed an attention mechanism that enables GANs to generate fine-grained images via word-level conditioning input.
However, all of their GANs are conditioned on the language descriptions without disentangling the semantic information under the expression variants. 
In our work, we focus on disentangling the semantic-related concepts to maintain the generation consistency from complex and various natural language descriptions as well as the details for text-to-image generation.

% % \vspace{0.1cm}
\noindent\textbf{Conditional Batch Normalization (CBN).}
Batch normalization (BN) is a widely used technique to improve neural network training by normalizing activations throughout the network with respect to each mini-batch. 
BN has been shown to accelerate training and improve generalization by reducing covariate shift throughout the network~\cite{ioffe2015batchnorm}.
Dumoulin \etal \cite{dumoulin2017learned} proposed a conditional instance normalization layer that learns the modulation parameters with the conditional cues. 
These parameters are used to control the behavior of the main network for the tasks such as image stylization~\cite{huang2017arbitrary}, visual reasoning~\cite{perez2017learning}, video segmentation~\cite{yang2018efficient}, question answering~\cite{de2017modulating} and \etc.
In our work, conditional batch normalization is firstly adopted for visual feature generation and the semantic-conditioned batch normalization layers enhance the visual-semantic embedding and the proposed layers are implemented in the generators of GANs for the purpose of the efficient visual generation based on the linguistic conditions. 
%

%=========== fig: module ==================
\begin{figure*}[t]
\centering
\includegraphics[width=0.85\linewidth]{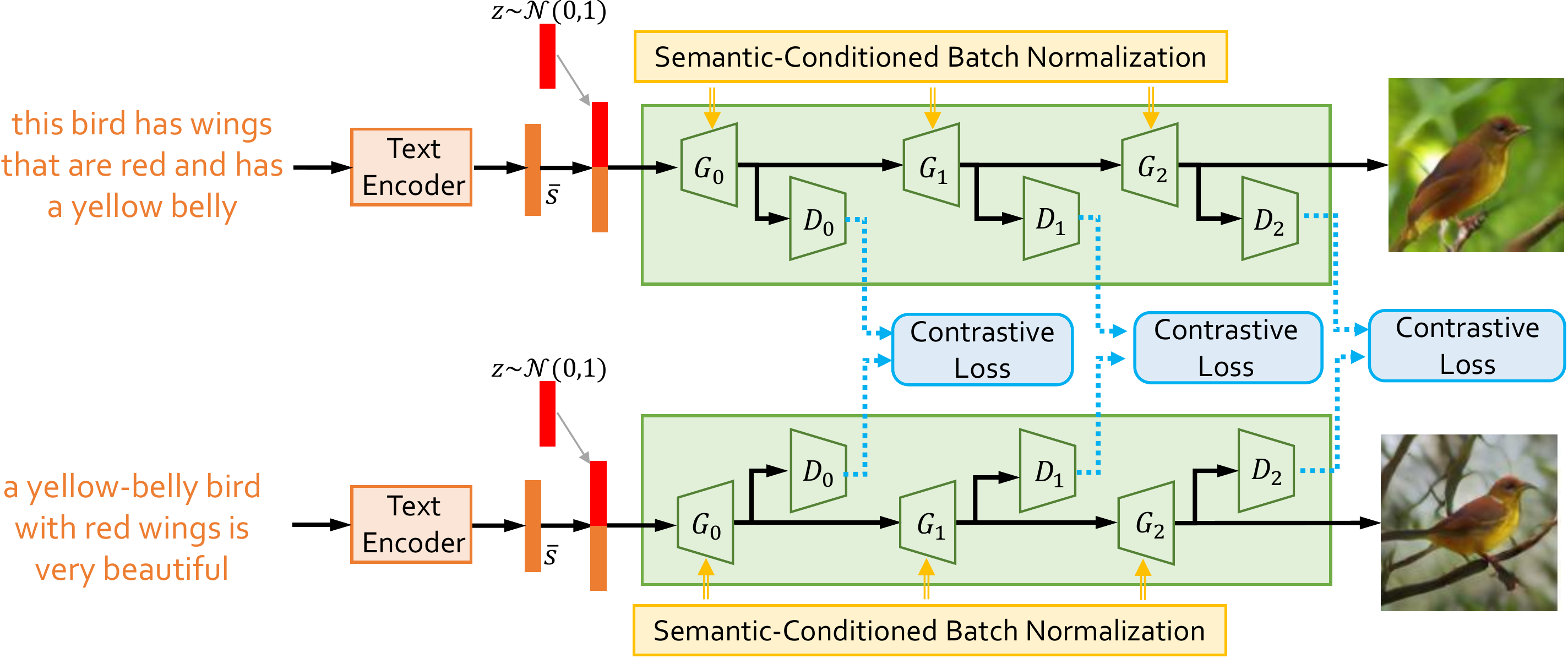}
% % \vspace{-0.5cm}
\caption{
The architecture of SD-GAN. The robust semantic-related text-to-image generation is optimized by contrastive losses based on a Siamese structure. The Semantic-Conditioned Batch Normalization (SCBN) is introduced to further retain the unique semantic diversities from text and embed the visual features modulated to the textual cues.
}
\label{fig:pipeline}
% \vspace{-0.25cm}
\end{figure*}
%=========== fig: module ==================

\section{Semantics Disentangling Generative Adversarial Network (SD-GAN)}
\label{sec:sd-gan}
% \vspace{-0.15cm}

In this paper, we propose a new cross-modal generation network named as Semantics Disentangling Generative Adversarial Network (SD-GAN) for text-to-image generation, as shown in Fig.~\ref{fig:pipeline}. It aims at distilling the \textit{semantic commons} from texts for image generation consistency and meanwhile retaining the \textit{semantic diversities \& details} for fine-grained image generation:
(1) Taking the advantages of Siamese structure, the generated images are not only based on the input description at the current branch, but also influenced by the description at the other branch. In other words, the Siamese structure distills the common semantics from texts to handle the generation deviation under the expression variants.
(2) To generate fine-grained visual patterns, the model also needs to retain the detailed and diverse semantics of the input texts. We modulate neural activations with linguistic cues by the proposed \textit{Semantic-Conditioned Batch Normalization} (SCBN), which will be introduced in Sec.~\ref{subsec:cbn}.

\subsection{Siamese Structure with Contrastive Losses}
\label{subsec:siamese}
% \vspace{-0.15cm}

Although existing methods~\cite{zhang2017stackgan,xu2017attngan} achieved excellent performances on high-resolution image generation, the generation deviations from language expression variants still pose great challenges for the text-semantic image generation.
To address the issues, the proposed SD-GAN adopts a Siamese structure for distilling textual semantic information for the cross-domain generation.
The contrastive loss is adopted for minimizing the distance of the fake images generated from two descriptions of the same groundtruth image while maximizing those of different groundtruth images.
During the training stage, the generated image is influenced by the texts from both two branches.

For constructing the backbone architecture for each Siamese branch, we adopt the sequential stacked generator-discriminator modules used in most previous works~\cite{zhang2017stackgan,xu2017attngan,hong2018inferring}. As shown in Fig.~\ref{fig:pipeline}, it consists of 
1) a text encoder \texttt{E} (in orange) for text feature extracting from descriptions, and 2) hierarchical generative adversarial subnets (in green) for image generation which contains a bunch of generators, \ie, $\texttt{G}_{0}, \texttt{G}_{1}, \texttt{G}_{2}$, and the corresponding adversarial discriminators, \ie, $\texttt{D}_{0}, \texttt{D}_{1}, \texttt{D}_{2}$.

% % \vspace{0.05cm}
\noindent\textbf{Text Encoder.}
The input of each branch is a sentence of natural language description.
The text encoder \texttt{E} aims at learning the feature representations from the natural language descriptions and following ~\cite{zhang2017stackgan,zhang2017stackgan++,xu2017attngan}, we adopt a bi-directional Long Short-Term Memory (LSTM)~\cite{hochreiter1997long} that extracts semantic vectors from the text description.
Generally, in the bi-directional LSTM, the hidden states are utilized to represent the semantic meaning of a word in the sentence while the last hidden states are adopted as the global sentence vector, \ie, $w_{t}$ denotes the feature vector for the $t^{th}$ word and $\bar{s}$ denotes the sentence feature vector.

\begin{figure}[t]
\centering
\includegraphics[width=0.90\linewidth]{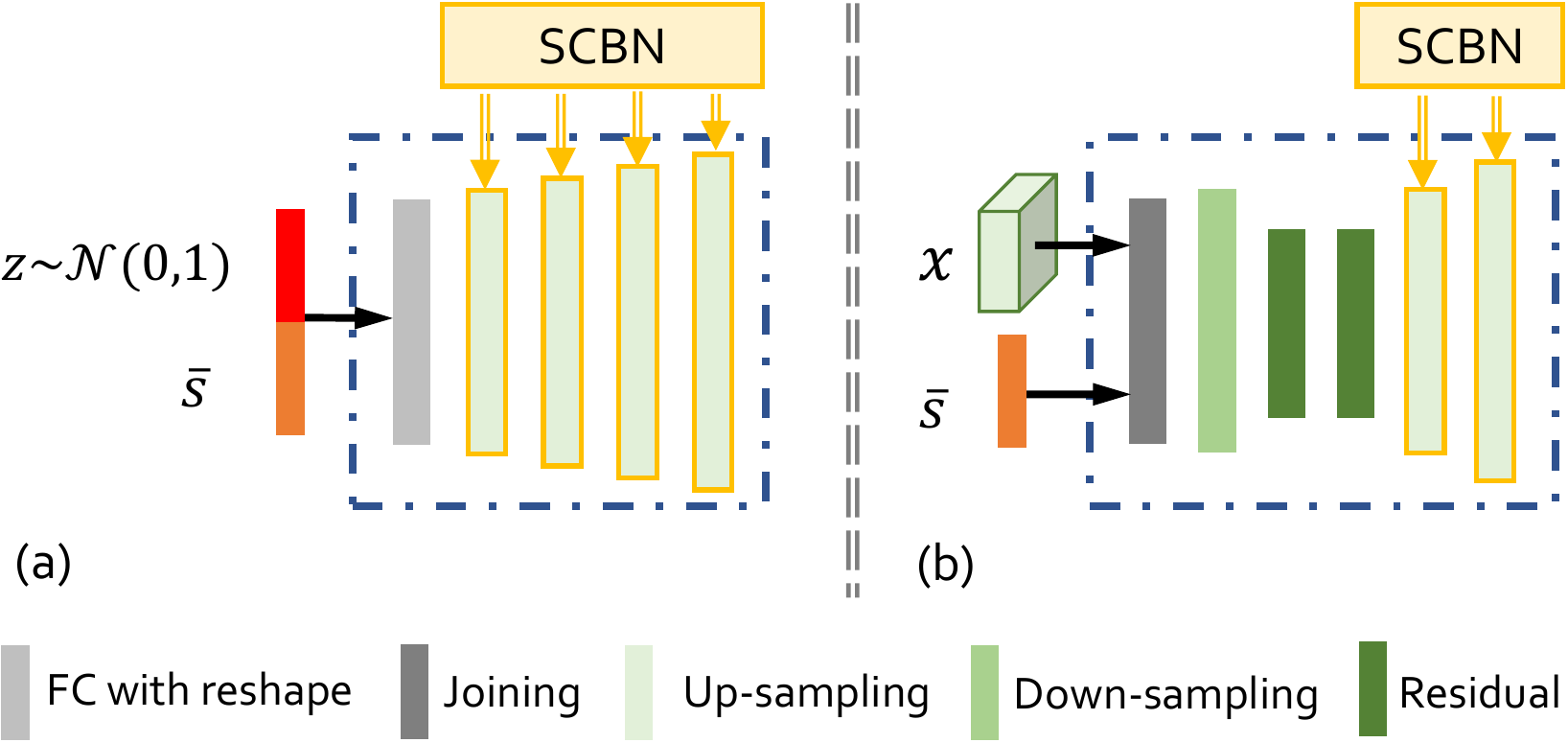}
% % \vspace{-0.5cm}
\caption{Illustration of the generators in the proposed SD-GAN: (a) $\texttt{G}_{0}$, the generator at the initial stage from the linguistic to vision, (b) $\texttt{G}_{1}/\texttt{G}_{2}$, the generator at the second/third stage for generating higher-resolution images based on generated visual features at the former stage. The SCBNs operate at the end of each up-sampling layer. 
}
\label{fig:generators}
% \vspace{-0.25cm}
\end{figure}

% % \vspace{0.05cm}
\noindent\textbf{Hierarchical Generative Adversarial Networks.}
Inspired by~\cite{zhang2017stackgan,xu2017attngan,hong2018inferring,zhang2017stackgan++}, we adopt hierarchical stages from low-resolution to high-resolution for the photo-realistic image generation.  
Given the sentence feature $\bar{s}$ from the text encoder \texttt{E} and a noise vector $z$ sampled from a standard normal distribution, the low resolution ($64 \times{64}$) image is generated at the initial stage, as shown in Fig.~\ref{fig:generators}~(a).
( The SCBN layer in Fig.~\ref{fig:generators} will be introduced in Sec.~\ref{subsec:cbn}. )
The following stage uses the output of the former stage as well as the sentence feature $\bar{s}$ to generate the image with higher-resolution, as shown in Fig.~\ref{fig:generators}~(b).
At each stage, the generator is followed by a discriminator that distinguishes whether the image is real or fake. These discriminators $\texttt{D}_{0}, \texttt{D}_{1}, \texttt{D}_{2}$ are independent for extracting the visual features and will not share parameters. 
% 

%=========== fig: module ==================
%=========== fig: module ==================

% % \vspace{0.05cm}
\noindent\textbf{Contrastive Loss.}
In our work, the purpose of the proposed Siamese structure is to enhance the generation consistency regardless of the expression variants of the input descriptions during the training procedure. 
We input two different text descriptions to the two branches of the Siamese structure respectively.
If the visual features generated from two branches are textual semantic-aware, the two generated images should be similar (\ie~with a small distance). Otherwise, the two generated images should be different (\ie~with a large distance).
To this end, we adopt the contrastive loss to distill the semantic information from the input pair of descriptions.

The contrastive loss is firstly introduced in \cite{hadsell2006contrastiveloss} and the loss function is formulated as 
\begin{equation}
% \vspace{-0.1cm}
L_{c} = \frac{1}{2N} \sum_{n=1}^{N} y \cdot d^{2} + (1 - y) \max(\varepsilon-d, 0)^{2},
\label{eqa:contrastive_loss}
\end{equation}
where $ d = \| v_1 - v_2 \|_{2}$ is the distance between the visual feature vectors $v_1$  and $v_2$ from the two Siamese branches respectively, and $y$ is a flag to mark whether the input descriptions are from the same image or not, \ie, $1$ for the same and $0$ for different. The hyper-parameter $N$ is the length of the feature vector and its value is set as $256$ empirically in the experiments. 
The hyper-parameter $\varepsilon$ is used to balance the distance value when $y=0$ and its value is set as $1.0$ in the experiments.

With the contrastive loss, the Siamese structure is optimized by minimizing the distance between the generated images from the descriptions of the same image and maximizing the distance of those generated from the descriptions of different images.
Note that due to the input noises, even though the input descriptions are exactly the same, the generated images might be different more or less in appearance, \eg, pose, background and \etc.
To avoid collapsed nonsensical mode in the visualization (\ie, the generated images are too close in appearance), the distance of their feature vectors are not required to be ``zero''.
Therefore, we modify the Eq.~\ref{eqa:contrastive_loss} as 
\begin{equation}
% \vspace{-0.1cm}
L_{c} = \frac{1}{2N} \sum_{n=1}^{N} y \max(d, \alpha)^{2} + (1 - y) \max(\varepsilon-d, 0)^{2},
\label{eqa:modified_contrastive_loss}
\end{equation}
where $\alpha$ is a hyper-parameter to avoid the fake images generated too closely even though the input two descriptions are from the same image. We set $\alpha=0.1$ in the experiments.

\subsection{Semantic-Conditioned Batch Normalization (SCBN)}
\label{subsec:cbn}
% \vspace{-0.15cm}

In this work, we consider the linguistic concepts as the kernels of visual representations for cross-domain generation from linguistic to vision.
Inspired by the instance normalization in the existing works~\cite{huang2017arbitrary,de2017modulating,yang2018efficient}, 
we modulate the conditional batch normalization with the linguistic cues from the natural language descriptions, defined as \textit{Semantic-Conditioned Batch Normalization} (SCBN).
The purpose of SCBN is to reinforce the visual-semantic embedding in the feature maps of the generative networks. It enables the linguistic embedding to manipulate the visual feature maps by scaling them up or down, negating them, or shutting them off, \etc.
It complements to the Siamese structure introduced in Sec.~\ref{subsec:siamese} which only focuses on distilling semantic commons but ignore the unique semantic diversities in the text.

% % \vspace{0.1cm}
\noindent\textbf{Batch Norm -}
Given an input batch $x \in \mathbb{R}^{N \times C \times H \times W}$, BN normalizes the mean and standard deviation for each individual feature channel as
\begin{equation}
% \vspace{-0.1cm}
\texttt{BN}(x) = \gamma \cdot \frac{x - \mu(x)}{\sigma(x)} + \beta,
\label{eqa:bn}
\end{equation}
where $\gamma, \beta \in \mathbb{R}^{C}$ are affine parameters learned from data, and $\mu(x), \sigma(x) \in \mathbb{R}^{C}$ are the mean and standard deviation which are computed across the dimension of batch and spatial independently for each feature channel.

% % \vspace{0.1cm}
\noindent\textbf{Conditional Batch Norm -}
Apart from learning a single set of affine parameters $\gamma$ and $\beta$, Dumoulin \etal \cite{dumoulin2017learned} proposed the Conditional Batch Normalization (CBN) that learns the modulation parameters $\gamma_{c}$ and $\beta_{c}$ with the conditional cues $c$. 
The CBN module is a special case of the more general scale-and-shift operation on feature maps.
The modified normalization function is formatted as
\begin{equation}
% \vspace{-0.1cm}
\texttt{BN}(x | c) = (\gamma + \gamma_{c}) \cdot \frac{x - \mu(x)}{\sigma(x)} + (\beta + \beta_{c}).
\label{eqa:cbn}
\end{equation}

%=========== fig: cbn ==================
\begin{figure}[t]
\centering
\includegraphics[width=0.9\linewidth]{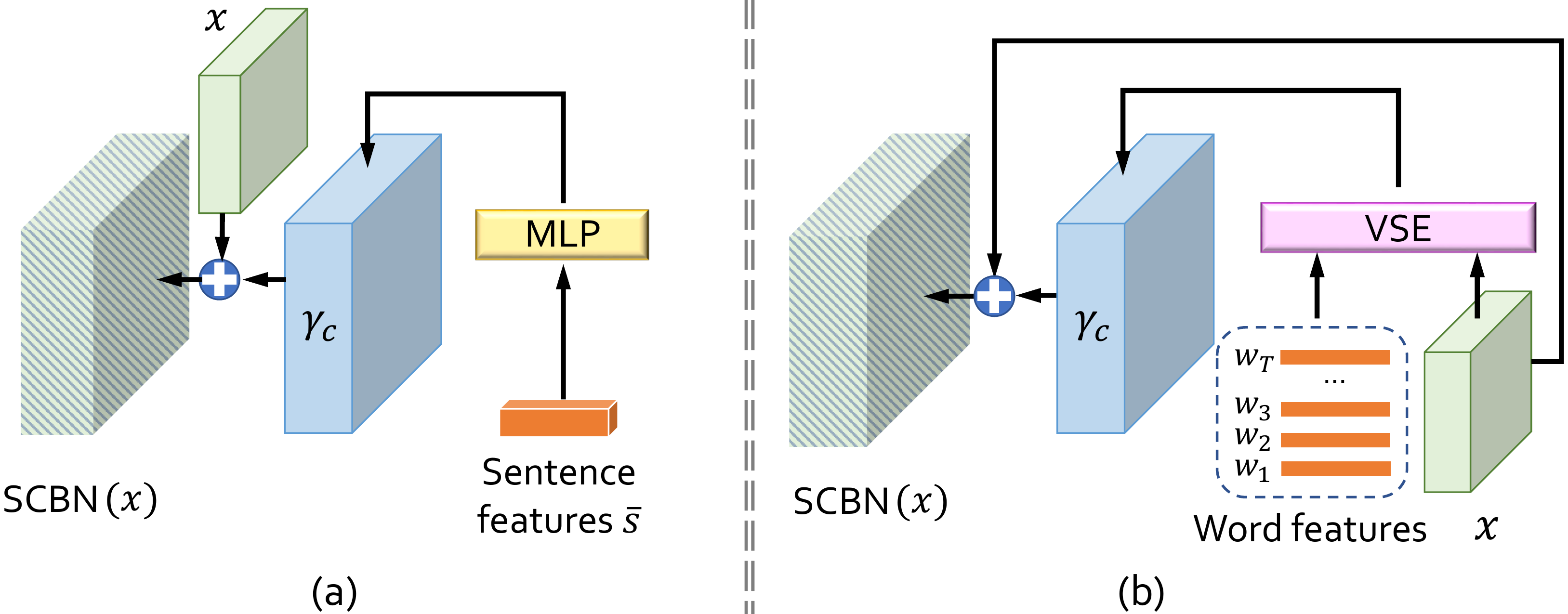}
% % \vspace{-0.5cm}
\caption{Semantic-conditioned batch normalization (SCBN) with (a) sentence-level cues that consists of a one-hidden-layer MLP to extract modulation parameters from the sentence feature vector; and (b) word-level cues that uses VSE module to fuse the visual features and word features. Note that the illustration only takes $\gamma_c$ as the example and the implementation for $\beta_c$ is alike.
}
\label{fig:scbn}
% \vspace{-0.25cm}
\end{figure}
%=========== fig: cbn ==================

% % \vspace{0.1cm}
\noindent\textbf{Semantic-Conditioned Batch Normalization - }
To reinforce the visual-semantic embedding for the visual generation, we implement the proposed SCBN layers in the generators, as shown in Fig.~\ref{fig:generators}.
Firstly, we recap the text encoder (\ie, bi-directional LSTM) to obtain the linguistic features from the input description.
Denote the linguistic features of the $t^{th}$ word as $w_t$. The last hidden states are adopted as the global sentence vector $\bar{s}$.
Therefore, the linguistic cues for SCBN can be obtained from two aspects, \ie, sentence-level and word-level.

% % \vspace{0.1cm}
\noindent (1) \textit{Sentence-level Cues.}
\label{subsubsec:scbn}
In order to embed the sentence feature, we adopt a one-hidden-layer multi-layer perceptron (MLP) to extract modulation parameters $\gamma_{c}$ and $\beta_{c}$ respectively from the sentence feature vector $\bar{s}$ of the input description, as shown in Fig.~\ref{fig:scbn}~(a). 
\begin{equation}
% \vspace{-0.1cm}
\gamma_{c} = f_{\gamma}(\bar{s}), \beta_{c} = f_{\beta}(\bar{s}),
\label{eqa:scbn}
\end{equation}
where $f_{\gamma}(\cdot)$ and $f_{\beta}(\cdot)$ denote the one-hidden-layer MLPs for $\gamma_c$ and $\beta_c$ respectively. 
Then we extend the dimension of $f_{\gamma}(\bar{s})$ and $f_{\beta}(\bar{s})$ to the same size as $x$ for embedding the linguistic cues and visual features with Eq.~\ref{eqa:cbn}.
Then the instance sentence features modulate the neural activations of the generated visual features by channel-wise.

\noindent (2) \textit{Word-level Cues.}
\label{subsubsec:wcbn}
Denote $\mathcal{W} = \{w_t\}_{t=1}^{T}\in\mathbb{R}^{D\times T}$ as the set of word features, where $w_t$ is the feature of the $t$-th word, and $\mathcal{X}\in\mathbb{R}^{C\times L}$ as the visual features where $C$ is the channel size and $L=W\times H$.
Inspired by~\cite{yu2017multi, NIPS2013_5204,faghri2018vse++,xu2017attngan}, the visual-semantic embedding (VSE) module is adopted for mutual fusion of word features and visual features, as shown in Fig.~\ref{fig:scbn}~(b).
We first use a perception layer (\ie, $f(w_t)$) to match the dimension of textual features and visual features. Then the VSE vector $\texttt{vse}_j$ is computed for each sub-region $j$ of the image based on its embedded features $v_j$ which is a dynamic representation of word vectors $\{w_{t}\}_{t=1}^{T}$ relevant to its visual feature $v_{j}$. 
\begin{equation}
% \vspace{-0.2cm}
\texttt{vse}_{j} = \sum_{t=0}^{T-1}\sigma(v_{j}^\top\cdot f(w_t)) f(w_{t}),
\label{eqa:wcbn}
\end{equation}
where $\sigma(v_{j}^\top\cdot f(w_t))$ indicates the visual-semantic embedding weight of $t^{th}$ word vector $w_t$ for the $j^{th}$ sub-region $v_j$ of visual feature maps, similar as the dot-product similarity of cross correlation. $\sigma(\cdot)$ is the \texttt{softmax} function in the experiments.
We then adopt two $\texttt{conv}\_{1\times1}$ layers for computing the word-level modulation parameters $\gamma_c$ and $\beta_c$ respectively from the VSE matrix.

\section{Experiments}
\label{sec:exps}
% \vspace{-0.15cm}

\subsection{Experiment Settings}
\label{subsec:exp_setting}
% \vspace{-0.15cm}

% % \vspace{0.05cm}
\noindent\textbf{Datasets.}
Following previous text-to-image methods \cite{xu2017attngan,zhang2017stackgan,zhang2017stackgan++}, our method is evaluated on CUB~\cite{wah2011cub200} and MS-COCO~\cite{lin2014microsoftcoco} datasets.
The CUB dataset contains $200$ bird species, it includes $11788$ images with $10$ language descriptions for each image.  
Following the settings in \cite{xu2017attngan,zhang2017stackgan,zhang2017stackgan++}, we split the CUB dataset into class-disjoint training and test sets, \ie, $8855$ images for training and $2933$ for test. All images in CUB dataset are preprocessed and cropped to ensure that bounding boxes of birds have greater-than-0.75 object-image size ratios.
The MS-COCO dataset is more challenging for text-to-image generation. It has a training set with $80k$ images and a validation set with $40k$ images. It has $5$ language descriptions for each image.

% % \vspace{0.1cm}
\noindent\textbf{Training Details.}
Apart from the contrastive losses introduced in Sec.~\ref{subsec:siamese}, the generator and the discriminator losses of the proposed SD-GAN follow those in~\cite{xu2017attngan} due to its excellent performance.
The text encoder and inception model for visual features used in visual-semantic embedding are pretrained by~\cite{xu2017attngan} and fixed during the end-to-end training.
\footnote{We also finetuned these models with the whole network, however the performance was not improved.}
The network parameters of the generator and discriminator are initialized randomly.
%

% % \vspace{0.1cm}
\noindent\textbf{Evaluation Details.}
It is not easy to evaluate the performance of the generative models. Following prior arts on text-to-image generation ~\cite{xu2017attngan,zhang2017stackgan,zhang2017stackgan++,hong2018inferring,zhang2018photographic,johnson2018scenegraph}, we apply the numerical assessment approach ``inception score''~\cite{salimans2016improved} for quantitative evaluation.
In our experiments, we directly use the pre-trained Inception model provided in \cite{zhang2017stackgan} to evaluate the performance on CUB and MS-COCO datasets.

Although the inception score has shown well correlated with human perception on visual quality~\cite{salimans2016improved}, it cannot tell whether the generated images are well conditioned on the text descriptions. Therefore, as a complementary, we also design a subject test to evaluate the generation performance.
We randomly select 50 text descriptions for each class in the CUB test set and 5000 text descriptions in the MS-COCO test set. 
Given the same descriptions, 50 users (not including any author) are asked to rank the results by different methods.
The average ratio ranked as the best by human users are calculated to evaluate the compared methods.

%=========== fig: cbn ==================
\begin{figure*}[t]
\centering
\includegraphics[width=\linewidth]{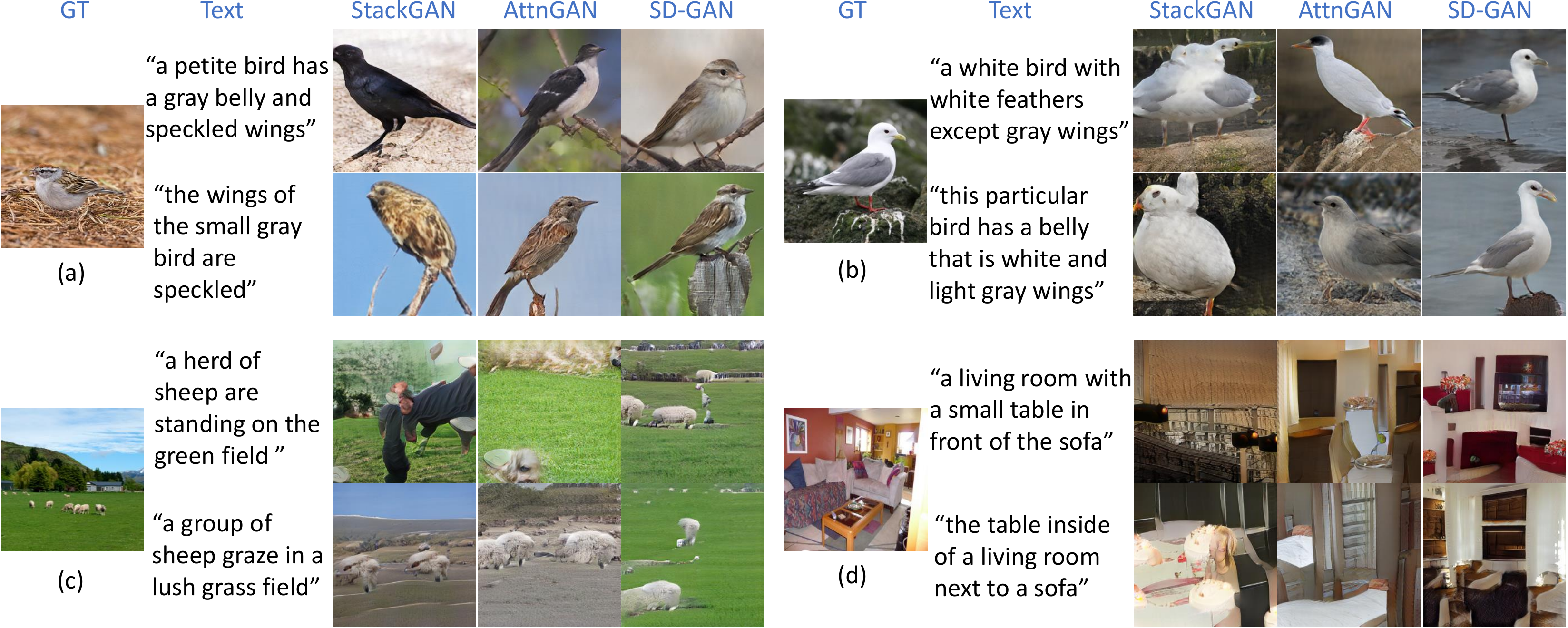}
% % \vspace{-0.5cm}
\caption{Qualitative examples of the proposed SD-GAN comparing with StackGAN~\cite{zhang2017stackgan} and AttnGAN~\cite{xu2017attngan} on CUB (top) and MS-COCO (bottom) test sets.
For each example, the images are generated by the methods based on two randomly-selected descriptions (Text) from the same ground-truth image (GT).
}
\label{fig:result_method}
% \vspace{-0.25cm}
\end{figure*}
%=========== fig: cbn ==================

\subsection{Comparing with the state-of-the-arts}
\label{subsec:state_of_the_arts}
% \vspace{-0.15cm}

We compare our results with the state-of-the-art text-to-image methods on CUB and MS-COCO dataset.
The inception scores for our proposed SD-GAN and other compared methods are listed in Tab.~\ref{tb:state_of_the_art}. 
On the CUB dataset, our SD-GAN achieves the inception score $4.67 \pm .09$ , which significantly outperforms the previous best method with an inception score $4.36 \pm .03$.
More impressively, our SD-GAN boosts the best reported inception score on the MS-COCO dataset from $ 25.89 \pm .47 $ to $35.69 \pm .50 $.
\footnote{
The inception score of CUB dataset is much lower than that of MS-COCO because the CUB dataset consists of fine-grained bird images while MS-COCO consists of images from more diverse scenarios. The generated images in MS-COCO is more suitable to be classified by the Inception model.}
The excellent performances on the datasets demonstrate the effectiveness of our proposed SD-GAN, thanks to the semantics-disentangling generation and visual-semantic embedding.

The results of subjective test are shown in Tab.~\ref{tb:human}.
We compared the proposed SD-GAN with the previous methods, \ie, StackGAN~\cite{zhang2017stackgan} and AttnGAN~\cite{xu2017attngan}.
When users are asked to rank images based on their relevance to input text, they choose the generated images by SD-GAN as the best mostly, wining about $70\%$ of the presented texts, much higher than others.
This is consistent with the improvements of inception score listed in Table~\ref{tb:state_of_the_art}. 
Furthermore, the qualitative results are shown in Fig.~\ref{fig:result_method}.
For each example, we compare the generation results from the descriptions of the same ground-truth image.
Due to the lacking of the word-level details, StackGAN fails to predict the important semantic structure of object and scene. 
Although AttnGAN adopts the attention mechanism to extract details from the text, it is difficult to generate the corresponding visual concepts under linguistic expression variants, \eg, \textit{gray wings of white bird} in Fig.~\ref{fig:result_method}(b), \textit{sheep on the grass} in Fig.~\ref{fig:result_method}(c),  and \etc. 
Comparing to them, the proposed SD-GAN generates more recognizable and semantically meaningful images based on the input texts.

%=========== tb: state_of_the_art ==================
\begin{table}[t]
\centering

\small
\begin{tabular}{R{3.3cm}|M{1.8cm}|M{1.8cm}} 
\hlinew{1.0pt}
Methods & CUB & MS-COCO \\
% \multirow{2}{*}{Methods} &\multicolumn{2}{c}{Dataset}\\
% & CUB & MS-COCO \\
\hline
GAN-INT-CLS~\cite{reed2016generative} & $2.88 \pm .04 $ & $7.88 \pm .07 $ \\

GAWWN ~\cite{reed2016learning} & $3.62 \pm .07 $ & -\\

StackGAN~\cite{zhang2017stackgan} & $3.70 \pm .04 $ & $8.45 \pm .03 $ \\ 

StackGAN++~\cite{zhang2017stackgan++} & $4.04 \pm .05 $ & - \\ 

PPGN~\cite{nguyen2017plug} & - & $9.58 \pm .21 $ \\ 

AttnGAN~\cite{xu2017attngan} & \underline{$4.36 \pm .03$} &  \underline{$ 25.89 \pm .47 $} \\

HDGAN~\cite{zhang2018photographic} & $ 4.15 \pm .05 $ & $11.86 \pm .18 $\\

Cascaded C4Synth~\cite{joseph2018c4synth} & $3.92 \pm .04 $& -\\

Recurrent C4Synth ~\cite{joseph2018c4synth} & $4.07 \pm .13 $ & - \\

LayoutSynthesis~\cite{hong2018inferring} & - & $11.46 \pm .09 $ \\

SceneGraph~\cite{johnson2018scenegraph} & - & $6.70 \pm .01 $ \\

\hline

SD-GAN & $\mathbf{4.67} \pm \mathbf{.09} $ & $\mathbf{35.69} \pm \mathbf{.50} $\\
% SD-GAN + AttnGAN & $\mathbf{4.67} \pm \mathbf{.09} $ & $\mathbf{38.69} \pm \mathbf{.56} $\\

\hlinew{1.0pt}
\end{tabular}

\caption{Quantitative results of the proposed method comparing with the state-of-the-arts on CUB and MS-COCO test sets. 
The bold results are the highest and the underline ones are the second highest.
}
% % \vspace{-0.25cm}
\label{tb:state_of_the_art}

\end{table}
%=========== tb: state_of_the_art ==================

%=========== tb: state_of_the_art ==================
\begin{table}[t]
\centering
\small
\begin{tabular}{R{2.3cm}|M{1.8cm}|M{1.8cm}} 
\hlinew{1.0pt}
Methods & CUB & MS-COCO \\

\hline

StackGAN~\cite{zhang2017stackgan} & $10.70 \%$ & $6.53 \% $ \\ 

AttnGAN~\cite{xu2017attngan} & $20.54 \%$ & $17.69 \% $ \\

SD-GAN & $68.76 \%$ & $75.78 \% $ \\

\hlinew{1.0pt}
\end{tabular}
\caption{Human evaluation results (ratio of 1st by human ranking) of SD-GAN comparing with StackGAN~\cite{zhang2017stackgan} and AttnGAN~\cite{xu2017attngan}. 
}
% % \vspace{-0.25cm}
\label{tb:human}

\end{table}
%=========== tb: state_of_the_art ==================

%=========== fig: cbn ==================
\begin{figure*}[t]
\centering
\includegraphics[width=\linewidth]{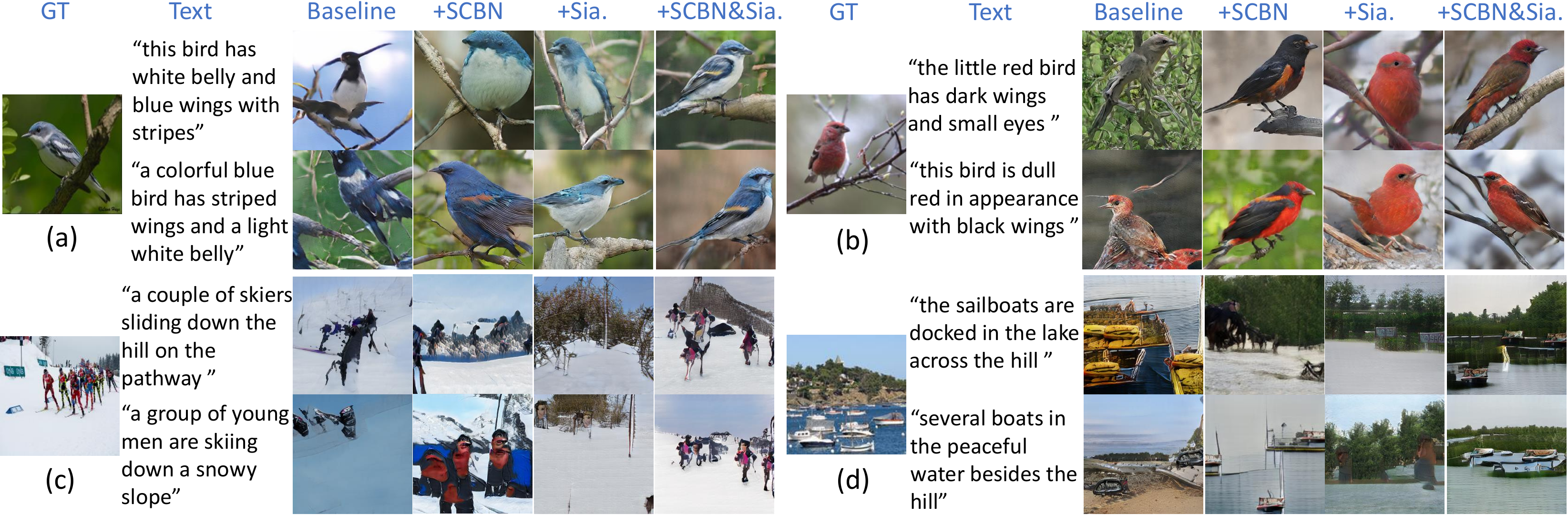}
% % \vspace{-0.5cm}
\caption{Image generation results of SD-GAN on CUB (top) and MS-COCO (bottom) test sets. For each sample, the images are generated by the methods based on two randomly-selected descriptions (Text) per ground-truth image (GT).
The results of of baseline (SD-GAN without SCBN\&Siamese) and its variants by adding the proposed SCBN and Siamese structure (Sia.) step by step. 
}
\label{fig:result_component}
% \vspace{-0.25cm}
\end{figure*}
%=========== fig: cbn ==================

% % \vspace{0.1cm}
\noindent \textbf{Transferable Siamese structure and SCBN.} Furthermore, we demonstrate the benefits of the proposed Siamese structure and SCBN for image generation by plugging them into the existing works.
Here we take the previous method, \ie, AttnGAN~\cite{xu2017attngan}, as the backbone because of its excellent performance. 
We compare three configurations, \ie, \textit{AttnGAN + Siamese}, \textit{AttnGAN + SCBN} and \textit{AttnGAN + Siamese + SCBN} under the same hyper-parameters for fair comparisons. 
As shown in Tab.~\ref{tb:transfer_learning}, the performance of AttnGAN is improved by a considerable margin on the inception score after applying the Siamese structure (\ie, \textit{AttnGAN + Siamese}).
The results again suggest the superiority of the proposed Siamese structure which is applied on AttnGAN.
AttnGAN with SCBNs (\ie, \textit{AttnGAN + SCBN}) achieves a better performance than AttnGAN as well. 
Note that the overall performance by adding both Siamese structure and SCBN (\ie, \textit{AttnGAN + Siamese + SCBN}) surpasses that of AttnGAN itself and achieves the approximate results with our proposed SD-GAN. 
%

%=========== tb: transfer_learning ==================
\begin{table}[t]
\centering

\small
\begin{tabular}{R{4.4cm}|M{1.3cm}|M{1.5cm}} 
\hlinew{1.0pt}
Methods & CUB & MS-COCO \\
\hline
AttnGAN~\cite{xu2017attngan} & {$4.36 \pm .03$} &  {$ 25.89 \pm .47 $} \\
\hline

AttnGAN~\cite{xu2017attngan} + Siamese & {$4.47 \pm .09$} &  {$ 29.77 \pm .51 $} \\

AttnGAN~\cite{xu2017attngan} + SCBN & {$4.48 \pm .08$} &  {$ 29.42 \pm .45 $} \\

AttnGAN~\cite{xu2017attngan} + Siamese + SCBN & $4.62 \pm .09 $ & $35.50 \pm .56$\\

% \hline
\hlinew{1.0pt}
\end{tabular}

\caption{Quantitative results of the combined models that incorporate the proposed Siamese structure and SCBN into the previous state-of-the-art architecture on CUB and MS-COCO test sets. 
}
% % \vspace{-0.25cm}
\label{tb:transfer_learning}

\end{table}
%=========== tb: transfer_learning ==================

\subsection{Component Analysis}
\label{subsec:ablation_study}
% \vspace{-0.15cm}

In this section, to evaluate the effectiveness of the proposed SCBN and Siamese structure with contrastive losses, we first quantitatively evaluate SD-GAN and its variants 
by removing each individual cue step by step, \ie, 
1) \textit{SD-GAN w/o SCBN} (Model $2$), SD-GAN without the proposed SCBNs, 
2) \textit{SD-GAN w/o Siamese} (Model $3$), SD-GAN without Siamese structure, 
3) \textit{SD-GAN w/o SCBN \& Siamese} (Model $4$), SD-GAN without the proposed SCBNs and Siamese structure, regarded as the baseline of SD-GAN.
The quantitative results are listed in Tab.\ref{tb:component}.

By comparing Model $3$ (with SCBNs) and Model $4$ (baseline) in Tab.~\ref{tb:component}, the proposed SCBN can help to enforce the visual-semantic embedding, which significantly improves the inception score from $4.11$ to $4.49$ on CUB and $23.76$ to $29.79$ on MS-COCO.
When adopting the Siamese structure (Model $2$) based on Model $4$, the inception score can achieve $4.51$ (versus $4.11$) on CUB dataset. 
By combining the proposed SCBNs and Siamese structure, Model $1$ obtains a significantly improvement and outperforms Model $3$ by improving the inception score from $4.49$ to $4.67$ on CUB and $29.79$ to $35.69$ on MS-COCO. 
The Siamese structure makes it possible to maintain the generation consistency and handle the generation deviation because of the input expression variations.
The comparisons demonstrate the superiority of the proposed SCBN and Siamese structure for text-to-image generation.

% =========== tb: ablation_study ==================
\begin{table}[t]
\centering
\small
\begin{tabular}{ M{0.5cm}|M{1.0cm} M{1.0cm}|M{1.5cm} M{1.7cm}} 
\hlinew{1.0pt}
  \multirow{2}{*}{ID} & \multicolumn{2}{c|}{Components} & \multirow{2}{*}{CUB} & \multirow{2}{*}{MS-COCO} \\
% Methods & CUB & MS-COCO \\
 & Siamese & SCBN &   & \\
\hline
 1 &  $\surd$ &$\surd$  & $\mathbf{4.67} \pm \mathbf{.09} $ & $\mathbf{35.69} \pm \mathbf{.50} $  \\
 2 & $\surd$ & - & $4.51 \pm .07 $ & $30.18 \pm .47 $ \\
 3 & - &$\surd$  & $4.49 \pm .06 $ & $29.79 \pm .61 $  \\
 4 & - & - & $4.11 \pm .04 $ & $ 23.76 \pm .40 $  \\

\hlinew{1.0pt}
\end{tabular}

\caption{Component Analysis of the SD-GAN. \textit{Siamese} indicates adopting the Siamese structure and \textit{SCBN} indicates using the proposed SCBN layer. The bold results are the best.
}
% \vspace{-0.25cm}
\label{tb:component}

\end{table}
% %=========== tb: ablation_study ==================

To better understand the effectiveness of the proposed modules, we visualize the generation results of SD-GAN and its variants.
As shown in Fig.~\ref{fig:result_component}, the baseline without Siamese structure and SCBN just sketches the primitive shape of objects lacking the exact descriptions. 
By adding the proposed SCBN (+SCBN), the models learn to rectify defects by embedding more linguistic details into the generation procedure, \eg ``blue wings'' in Fig.~\ref{fig:result_component}(a), but the generated birds belong to different categories in appearance due to the expression variants.
The model with Siamese structure (+Sia.) can generate similar images from different descriptions of the same image, but might lose the detailed semantic informations, \eg, ``black wings'' in Fig.~\ref{fig:result_component}(b).
By combining the Siamese structure and SCBN (+SCBN\&Sia.), the models can achieve visibly significant improvements.
On the challenging MS-COCO dataset, we have similar observations.
Although the generation is far from perfection, the generated images can still be recognized from the text semantics as shown in the bottom of Fig.~\ref{fig:result_component}.
Those observations demonstrate that the SD-GAN not only maintain the generation consistency but also contains the detailed semantics.

\begin{figure}[t]
\centering

\includegraphics[width=\linewidth]{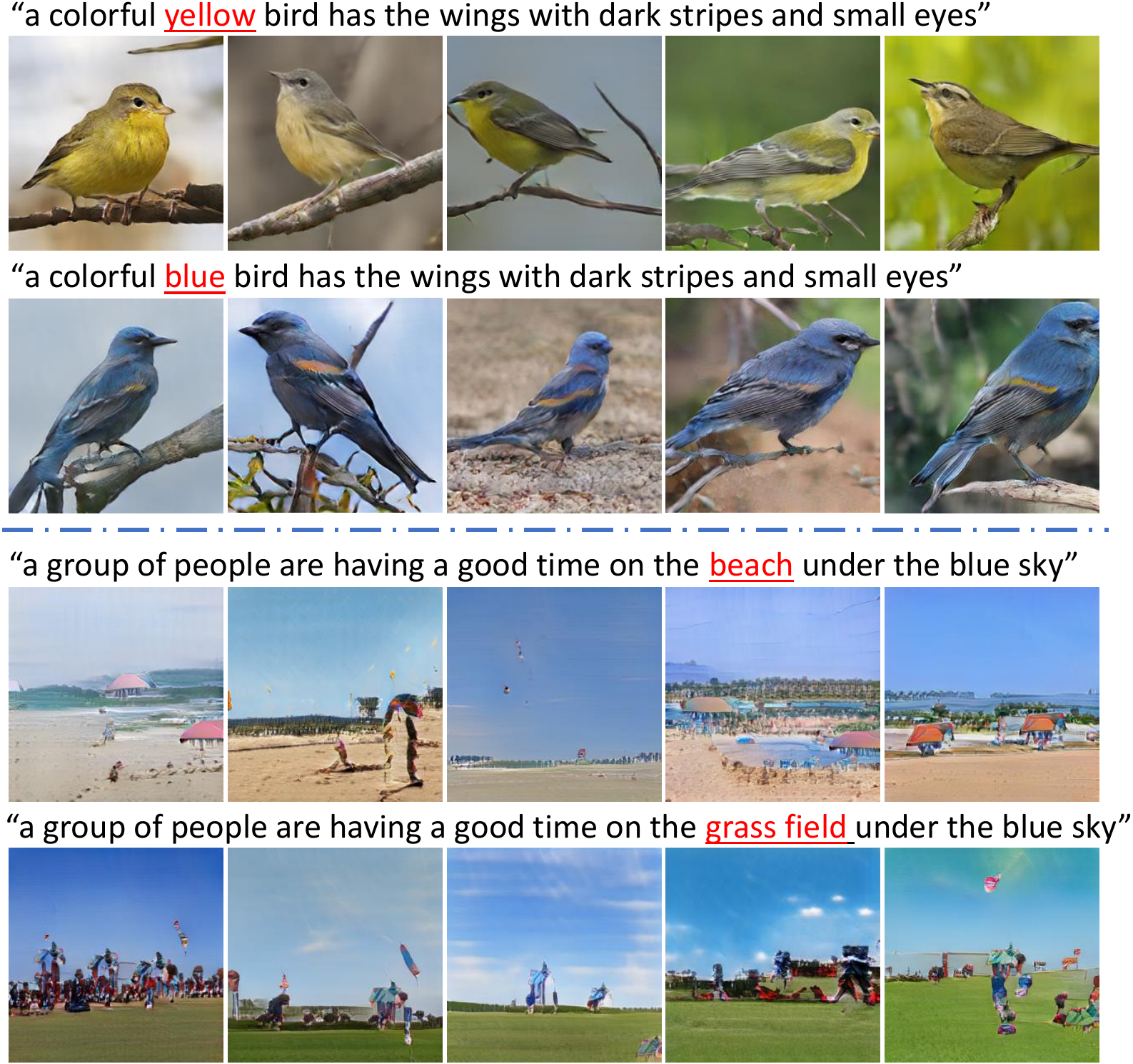}
% % \vspace{-0.5cm}
\caption{Examples of SD-GAN on the ability of catching subtle changes (underline word or phrase in red) of the text
descriptions on CUB (top) and MS-COCO (bottom) test sets.
}
\label{fig:result_mode}
% \vspace{-0.25cm}
\end{figure}

Furthermore, to evaluate the sensitivity of the proposed SD-GAN, we change just one word or phrase in the input text descriptions.
As shown in Fig.~\ref{fig:result_mode}, the generated images are modified according to the changes of the input texts, \eg, bird color (\textit{yellow} versus \textit{blue}) and image scene (\textit{beach} versus \textit{grass field}). 
It demonstrates the proposed SD-GAN retains the semantic diversities \& details from text and has the ability to catch subtle changes of the text descriptions.
On the other hand, there are no collapsed nonsensical mode in the visualization of the generated images.

%=========== fig: cbn ==================
%=========== fig: cbn ==================

% % \vspace{0.1cm}
\noindent\textbf{Contrastive Losses.}
The value of $\alpha$ in Eq.~\eqref{eqa:modified_contrastive_loss} is worth investigating because it can be used to find a trade-off between effectiveness of distilling semantic commons and retaining the semantic diversities from the descriptions of the same image.
We validate the value of $\alpha$ among $0.01, 0.05, 0.1$ and $0.2$ of SD-GAN. 
By comparing the results listed in Tab.~\ref{tb:component_siamese}, we adopt $\alpha$ as $0.1$ for further experiments as it has the best performances on both CUB and MS-COCO datasets.

Furthermore, we explore the effectiveness of contrastive losses at each stage by removing the contrastive loss stage by stage, \ie, 1) (D1, D2, D3) indicates the contrastive losses are implemented at all the stages as shown in Fig.~\ref{fig:pipeline}, 2) (D2, D3) indicates only at the last two stages and 3) (D3) indicates only at the last stage.
By comparing (D1, D2, D3) with (D2, D3) and (D3) in Tab.~\ref{tb:component_siamese}, the model with contrastive loss implemented at each stage (D1, D2, D3) achieves the best performances.
%

%=========== tb: ablation_study_siamese ==================
\begin{table}[t]
\centering
\small
\begin{tabular}{R{1.cm} | R{1.8cm}|M{1.3cm} M{2.0cm}} 
\hlinew{1.0pt}

\multicolumn{2}{c|}{Methods} & CUB & MS-COCO  \\
\hline

 \multirow{4}{*}{$\alpha$} & 0.01 & $ 4.50 \pm .08 $ & $32.53 \pm .77 $ \\
  & 0.05 & $ 4.55 \pm .10 $ & $33.18 \pm .62 $ \\
 & 0.1 & $ 4.67 \pm .09 $ & $35.69 \pm .50 $ \\
 & 0.2 & $ 4.49 \pm .07 $ & $31.74 \pm .91 $ \\
\hline
\multirow{3}{*}{position}& (D1, D2, D3)  & $4.67 \pm .09 $ & $35.69 \pm .50 $  \\
& (D2, D3) & $4.59 \pm .10 $ & $33.13 \pm .74 $ \\ 
& (D3) & $4.56 \pm .09 $ & $32.88 \pm .82 $ \\ 
% \hline

% \hline

\hlinew{1.0pt}
\end{tabular}
\caption{Ablation study on the contrastive loss. We compare the variants of SD-GAN with different values of hyper-parameter $\alpha$, \ie $0.01, 0.05, 0.1, 0.2 $. Then we compare the variants of SD-GAN by removing the contrastive loss at the individual stage.
}
% \vspace{-0.25cm}
\label{tb:component_siamese}

\end{table}
%=========== tb: ablation_study_siamese ==================

%=========== tb: ablation_study_scbn ==================
\begin{table}[t]
\centering

\small
\begin{tabular}{R{2.5cm}|M{1.5cm} M{2.0cm}} 
\hlinew{1.0pt}

Methods & CUB & MS-COCO \\
\hline

\textit{SCBN - sent} & $ 4.39 \pm .06 $ & $28.81 \pm 0.53 $  \\
\textit{SCBN - word} & $4.45 \pm .06 $ & $29.79 \pm 0.61 $  \\
\hline

\textit{BN - sent} & $4.19 \pm .05 $ & $24.18 \pm .56 $  \\
\textit{BN - word} & $4.23 \pm .05 $ & $25.34 \pm .79 $  \\
% \hline

\hlinew{1.0pt}
\end{tabular}
\caption{Ablation study on SCBN.
\textit{SCBN-sent} indicates using the SCBN layers conditioned on the sentence-level cues;
\textit{SCBN-word} indicates using the SCBN layers conditioned on the word-level cues;
\textit{BN-sent} indicates using BN layers and then concatenating sentence-level cues by channel-wise;
\textit{BN-word} indicates using BN layers and then concatenating word-level cues by channel-wise.
}
% \vspace{-0.25cm}
\label{tb:component_scbn}

\end{table}
%=========== tb: ablation_study_scbn ==================

% % \vspace{0.1cm}
\noindent\textbf{Semantic-Conditioned Batch Normalization (SCBN).}
To evaluate the benefits of the proposed SCBN layer, we compare the variants of the SCBN layers.
We conduct the experiments with the architecture of SD-GAN without Siamese structure due to the less computational cost during the training.
As introduced in Sec.~\ref{subsec:cbn}, the linguistic cues are from sentence-level and word-level.
Firstly, we compare the model with SCBN layer on sentence-level linguistic cues, \ie, \textit{SCBN - sent}, and that with word-level cues, \ie, \textit{SCBN - word}.
By comparing the results listed in Tab.~\ref{tb:component_scbn}, the SCBN layer with word-level cues outperforms that with sentence-level cues, \ie, $4.45$ versus $4.39$ on CUB dataset. 
The word-level features provide more details than the coarse sentence-level features and the visual-semantic embedding defined in Eq.~\eqref{eqa:wcbn} enables the visual modulation in the spatial configurations by the linguistic cues.

In addition, we replace the proposed SCBN layer with the general BN layer.
The linguistic cues are embedded into the visual feature maps as well by concatenating in channels directly after BN.
The BN layers with sentence-level and word-level cues are represented by \textit{BN - sent} and \textit{BN - word} respectively.
By comparing the results of \textit{SCBN - sent} versus \textit{BN - sent} and \textit{SCBN - word} versus \textit{BN - word} in Tab.~\ref{tb:component_scbn}, both of the SCBN layers outperform the corresponding BN layers in the experiments.
No doubt that the proposed SCBN is more efficient and powerful for embedding the linguistic cues into the generated vision.

\section{Conclusion} % (fold)
\label{sec:conclusion}
% \vspace{-0.25cm}
In this paper, we propose an innovative text-to-image generation framework,
named as Semantics Disentangling Generative Adversarial Networks (SD-GAN), that effectively exploit the semantics of the input text within the generation procedure.
The proposed SD-GAN adopts a Siamese structure to distills semantic commons from the linguistic descriptions, based on which the generated images can keep
generation consistency under expression variants.
Furthermore, to complement the Siamese mechanism that may lose unique semantic diversities, we design an enhanced visual-semantic embedding method by reformulating the batch normalization layer with the instance linguistic cues.
Extensive experiments demonstrate the respective effectiveness and significance of the proposed SD-GAN on the CUB dataset and the
challenging large-scale MS-COCO dataset.
% susionection concl (end)

\vspace{0.2cm}
\noindent\textbf {Acknowledgment}
% \subsubsection{Acknowledgment}
This work is supported in part by the National Natural Science Foundation of China (Grant No. 61371192), the Key Laboratory Foundation of the Chinese Academy of Sciences (CXJJ-17S044) and the Fundamental Research Funds for the Central Universities (WK2100330002, WK3480000005), in part
by SenseTime Group Limited, the General Research Fund sponsored by the Research Grants Council of Hong Kong (Nos. CUHK14213616, CUHK14206114, CUHK14205615, CUHK14203015, CUHK14239816, CUHK419412, CUHK14207-814, CUHK14208417, CUHK14202217), the Hong Kong Innovation and Technology Support Program (No.ITS/121/15FX).

{\small
\bibliographystyle{ieee_fullname}
\bibliography{ID462.bbl}
}

\clearpage

\section{Appendix}

%=========== fig: discriminators ==================
\begin{figure}[htp]
\centering
\includegraphics[width=\linewidth]{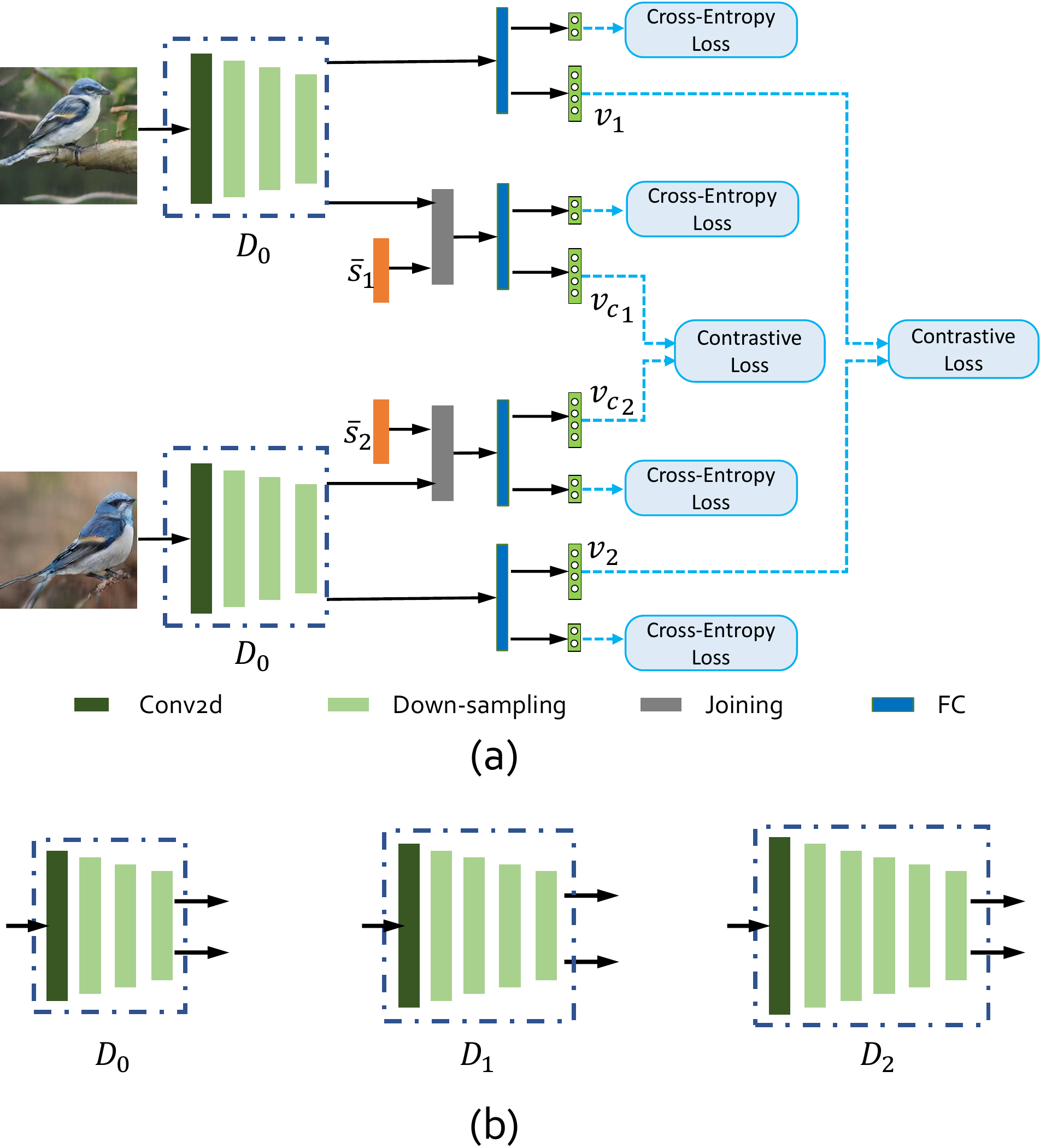}
% \vspace{-0.5cm}
\caption{
Network architecture of discriminators. 
}
\label{fig:discriminator}
\end{figure}
%=========== fig: discriminators ==================

\subsection{Architecture of Discriminators}
\label{sec:network}

As described in Sec.3 in the main paper, at each branch of the Siamese structure, we adopt hierarchical stages from low-resolution to high-resolution for the photo-realistic image generation.
At each stage, the generator is followed by a discriminator that distinguishes whether the image is real or fake.
%
% The network architecture of discriminators are shown in Fig.~\ref{fig:discriminator}.
%
As shown in Fig.~\ref{fig:discriminator}~(a), the input image is processed by several convolutional layers (in dot-line bounding box) for extracting the visual features. 
The visual features are fed into two branches, where each branch has two outputs, \ie, a classification vector for cross-entropy loss (1 for real, 0 for fake) and a feature vector for contrastive loss. 
Differs to the first branch, the second branch has its input as a concatenation of the sentence-level feature $\bar{s}$ and the visual feature map, following~\cite{zhang2017stackgan,xu2017attngan,zhang2017stackgan++}.
The contrastive loss (Eq.2 in the main paper) is calculated as follows,
\begin{equation}
% \vspace{-0.1cm}
\begin{aligned}
L_c = \frac{1}{2N} \sum_{n=1}^{N} y \max(d, \alpha)^{2} + (1 - y) \max(\varepsilon-d, 0)^{2} + \\
\frac{1}{2N} \sum_{n=1}^{N} y \max(d_c, \alpha)^{2} + (1 - y) \max(\varepsilon-d_c, 0)^{2},
\end{aligned}
\label{eqa:modified_contrastive_loss}
\end{equation}
where $ d = \| v_1 - v_2 \|_{2}$ , $ d_c = \| v_{c_1} - v_{c_2} \|_{2}$ is the distance between the visual feature vectors from the two Siamese branches respectively.

The discriminators $D_{0}, D_{1}, D_{2}$ in Fig.\ref{fig:discriminator}~(b) have similar structures. To obtain the output of each discriminator with the same size, the discriminator is constructed with different number of down-sampling layers. These discriminators are independent and will not share the parameters.

\subsection{More Results}
\label{sec:exp2}

\begin{figure*}[t]
\centering
\includegraphics[width=\linewidth]{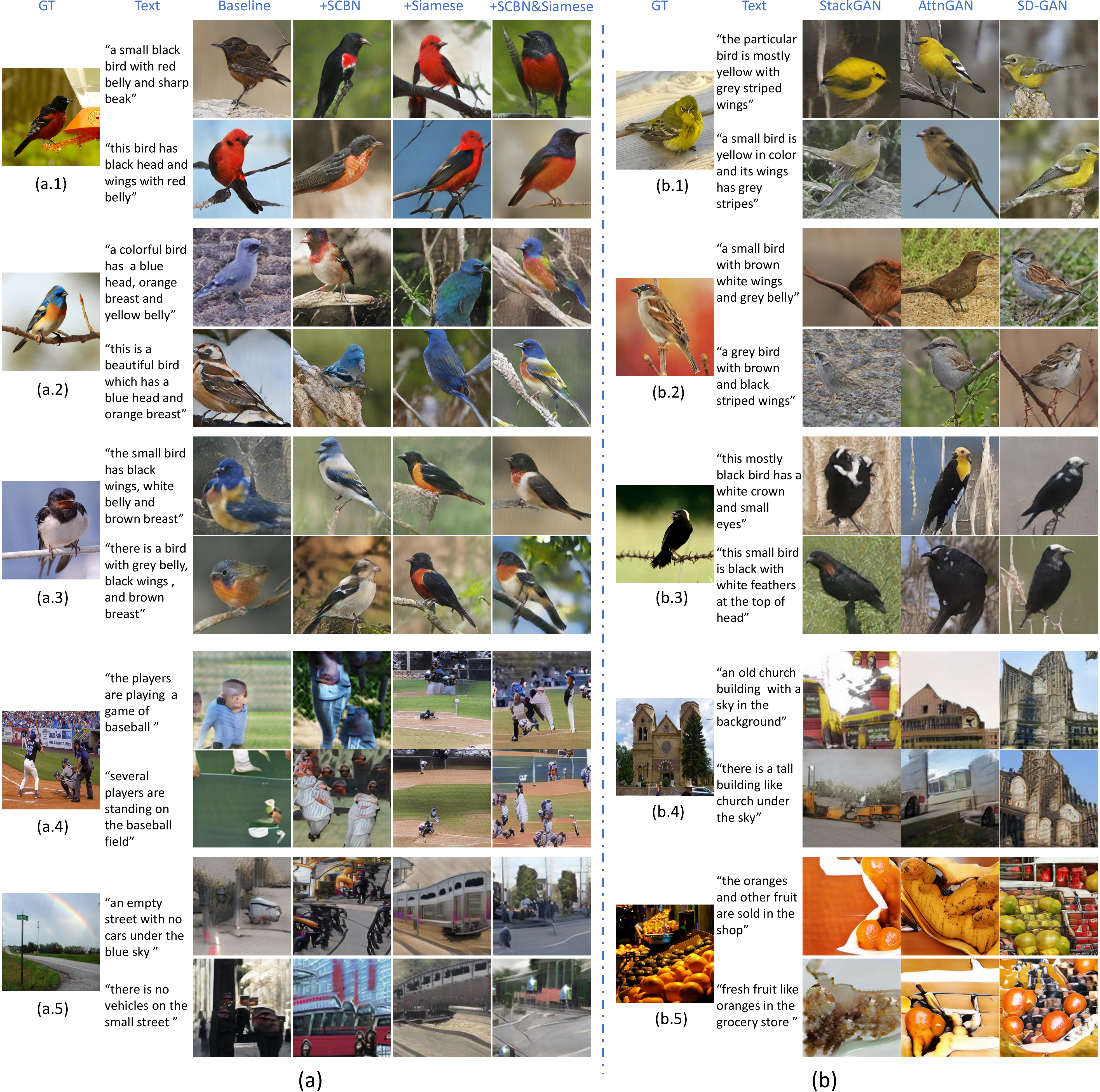}
% \vspace{-0.5cm}
\caption{
Additional qualitative resutls of the proposed SD-GAN. 
For each example, the images are generated by the methods based on two randomly-selected descriptions (Text) per ground-truth image (GT). 
(a) The results of baseline (SD-GAN without SCBN \& Siamese) and its variants by adding the proposed SCBN and Siamese structure step by step.
(b) The results of SD-GAN comparing with the state-of-the-art methods, \ie, StackGAN~\cite{zhang2017stackgan} and AttnGAN~\cite{xu2017attngan}.
}
\label{fig:additional_results}
\end{figure*}

\noindent \textbf{Additional qualitative results of SD-GAN.}
The additional qualitative comparisons are visualized in Fig.~\ref{fig:additional_results}: (1) we compare the results between different module configurations of the proposed SD-GAN, and (2) we show the excellent performance of SD-GAN, compared with the state-of-the-art methods, \ie, StackGAN~\cite{zhang2017stackgan} and AttnGAN~\cite{xu2017attngan}. The details are depicted in Sec.4.2 and Sec.4.3 in the main paper.

\begin{figure*}[t]
\centering
\includegraphics[width=1\linewidth]{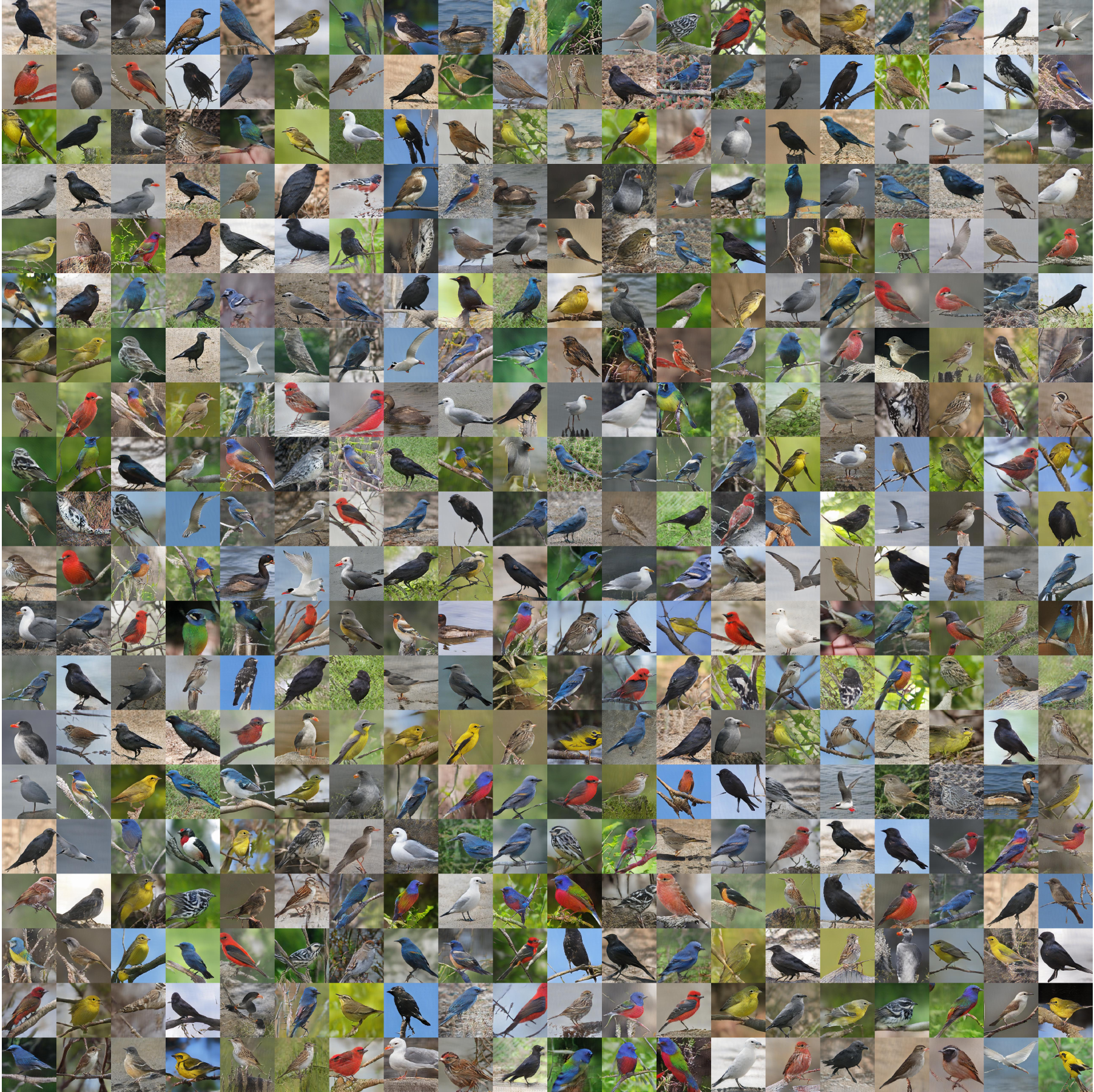}
% \vspace{-0.5cm}
\caption{Generated images randomly-sampled from CUB dataset (Please zoom in to view more details).
}
\label{fig:compressed_bird}
\end{figure*}

\begin{figure*}[t]
\centering
\includegraphics[width=\linewidth]{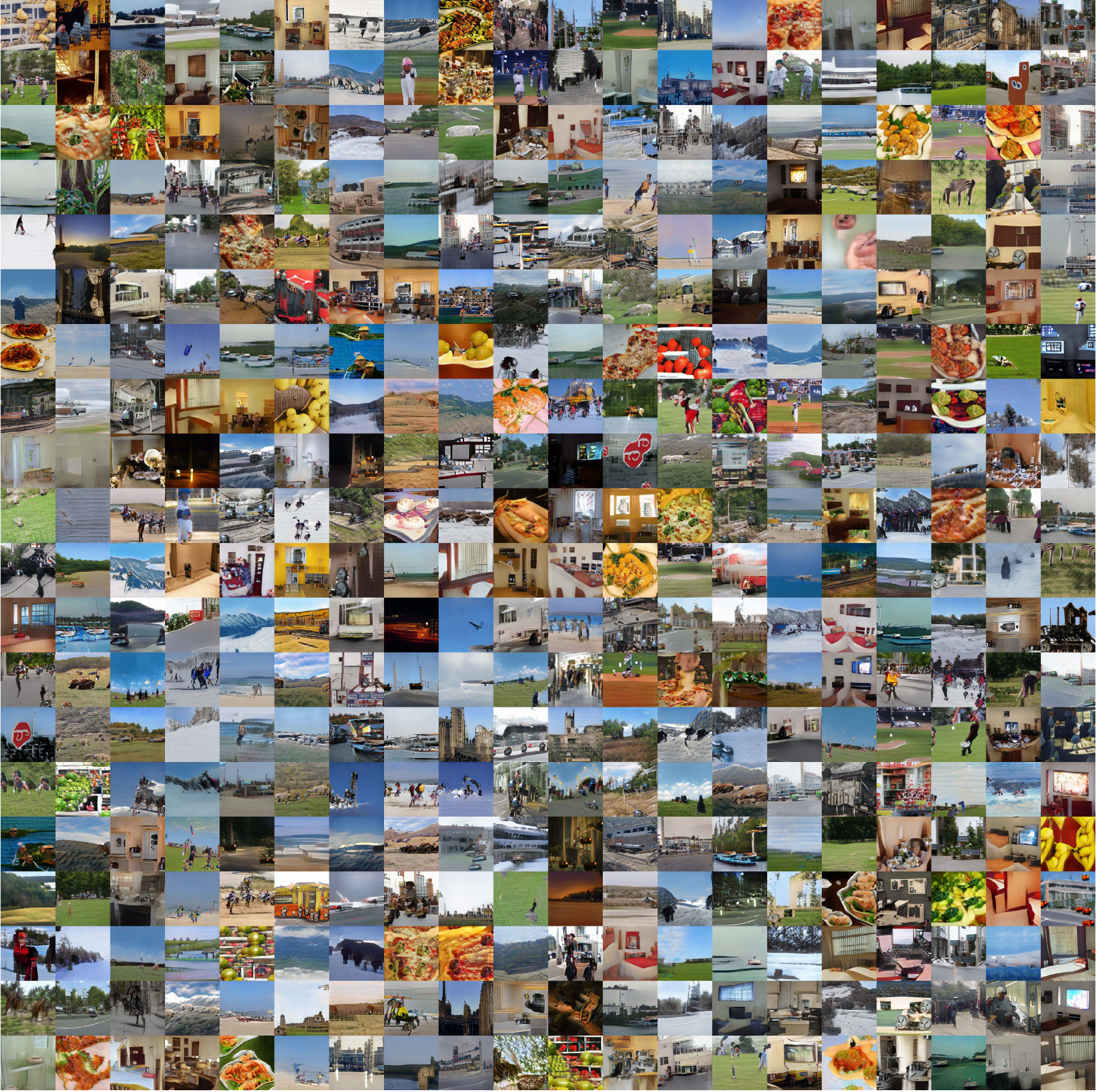}
% \vspace{-0.5cm}
\caption{Generated images randomly-sampled from MS-COCO dataset (Please zoom in to view more details).
}
\label{fig:compressed_coco}
\end{figure*}

\noindent \textbf{More generated results.}
When visualizing a large number of generated images by the proposed SD-GAN, we do not observe obvious nonsensical modes on both CUB and MS-COCO datasets.
Since the limited size of the supplementary material, here we only show $400$ images for each dataset, as shown Fig.~\ref{fig:compressed_bird} and Fig.\ref{fig:compressed_coco} respectively. These images are randomly selected, and the original resolution is $256 \times 256$ (Please zoom in to view more details).

%=========== fig: additional_results ==================
%=========== fig: additional_results ==================

%=========== fig: compressed_bird ==================
%=========== fig: compressed_bird ==================

%=========== fig: compressed_coco ==================
%=========== fig: compressed_coco ==================

\end{document}